\documentclass[sigconf,natbib=false]{acmart}
\AtBeginDocument{%
  }

\usepackage{tikz}
\usepackage{booktabs}
\usepackage{multirow}
\usepackage{amsthm}
\usepackage{todonotes}
\usepackage{tabularx,ragged2e,siunitx}
\usepackage{subcaption}
\usepackage{svg}
\usepackage{pgfplots}

\usepackage{algpseudocode}
\usepackage{tcolorbox}
\usepackage{balance}

\theoremstyle{definition}
\newtheorem{definition}{Definition}

\acmConference[SIGIR '48]{Research and Development in Information Retrieval}{July 13-18, 2025}{Padua, Italy}

\copyrightyear{2025}
\acmYear{2025}
\setcopyright{cc}
\setcctype{by}
\acmConference[SIGIR '25]{Proceedings of the 48th International ACM SIGIR Conference on Research and Development in Information Retrieval}{July 13--18, 2025}{Padua, Italy}
\acmBooktitle{Proceedings of the 48th International ACM SIGIR Conference on Research and Development in Information Retrieval (SIGIR '25), July 13--18, 2025, Padua, Italy}\acmDOI{10.1145/3726302.3729882}
\acmISBN{979-8-4007-1592-1/2025/07}

\RequirePackage[
  datamodel=acmdatamodel,
  style=acmnumeric,
]{biblatex}

\begin{document}

\title{A New HOPE: Domain-agnostic Automatic Evaluation~of~Text~Chunking}


\author{Henrik Brådland}
\orcid{0000-0002-1601-8884}
\affiliation{%
  \institution{Centre for Artificial Intelligence Research, University of Agder}
  \city{Kristiansand}
  \state{Agder}
  \country{Norway}
}
\affiliation{%
  \institution{Norkart AS}
  \city{Oslo}
  \country{Norway}
}
\email{henrik.bradland@uia.no}

\author{Morten Goodwin}
\orcid{0000-0001-6331-702X}
\affiliation{%
  \institution{Centre for Artificial Intelligence Research, University of Agder}
  \city{Kristiansand}
  \state{Agder}
  \country{Norway}
}

\author{Per-Arne Andersen}
\orcid{0000-0002-7742-4907}
\affiliation{%
  \institution{Centre for Artificial Intelligence Research, University of Agder}
  \city{Kristiansand}
  \state{Agder}
  \country{Norway}
}

\author{Alexander S. Nossum}
\orcid{0009-0005-1769-3194}
\affiliation{%
  \institution{Norkart AS}
  \city{Oslo}
  \country{Norway}
}

\author{Aditya Gupta}
\orcid{0000-0003-3128-2517}
\affiliation{%
  \institution{Centre for Artificial Intelligence Research, University of Agder}
  \city{Kristiansand}
  \state{Agder}
  \country{Norway}
}


\renewcommand{\shortauthors}{Brådland et al.}

\begin{abstract}
Document chunking fundamentally impacts Retrieval-Augmented Generation (RAG) by determining how source materials are segmented before indexing. Despite evidence that Large Language Models (LLMs) are sensitive to the layout and structure of retrieved data, there is currently no framework to analyze the impact of different chunking methods. In this paper, we introduce a novel methodology that defines essential characteristics of the chunking process at three levels: intrinsic passage properties, extrinsic passage properties, and passages-document coherence. We propose HOPE (Holistic Passage Evaluation), a domain-agnostic, automatic evaluation metric that quantifies and aggregates these characteristics. Our empirical evaluations across seven domains demonstrate that the HOPE metric correlates significantly ($\rho > 0.13$) with various RAG performance indicators, revealing contrasts between the importance of extrinsic and intrinsic properties of passages. Semantic independence between passages proves essential for system performance with a performance gain of up to $56.2\%$ in factual correctness and $21.1\%$ in answer correctness. On the contrary, traditional assumptions about maintaining concept unity within passages show minimal impact. These findings provide actionable insights for optimizing chunking strategies, thus improving RAG system design to produce more factually correct responses.
\end{abstract}



\keywords{Document Chunking, Passage Evaluation, Retrieval-Augmented Generation, Text Embedding, Natural Language Processing}


\maketitle

\section{Introduction}
\label{sec:introduction}

Document chunking, the process of dividing texts into coherent passages, plays a crucial role in RAG architectures \cite{Zhao2024RetrievalWisely}. Recent studies demonstrate that chunking strategies impact downstream task performance \cite{Yepes2024FinancialGeneration, Zhong2024Mix-of-Granularity:Generation, Barnett2024SevenSystem} and that LLMs exhibit sensitivity to passage format and structure \cite{Cuconasu2024TheSystems, Wu2024HowModels}. However, the field lacks formal definitions and evaluation methodologies that directly assess chunking quality independent of downstream applications. Current approaches primarily evaluate chunking through end-task performance, leaving unexplored fundamental questions about passage quality characteristics. To address this gap, this paper has the following main contributions:
\begin{itemize}
    \item A new, more open definition of chunking. 
    \item Three formalized principles for chunking.
    \item The HOPE metric, a holistic evaluation of passage quality.
    \item Experiments showcasing the impact of HOPE.
\end{itemize}


\subsection{Paper Outline}
In section~\ref{sec:introduction} the importance of chunking is stated, a definition is presented and principles for chunking are formulated. Section~\ref{sec:related_work} presents related work, while section~\ref{sec:method} introduces the methodology and the main contribution: The HOPE metric. Section~\ref{sec:results} describes the experimental setup and presents the empirical results, which are then discussed in section~\ref{sec:discussion}. The conclusion is given in section~\ref{sec:conclusion}.

\subsection{Defining Chunking}
Chunking traditionally refers to dividing documents into smaller text segments for Information Retrieval (IR) tasks~\cite{Wu2024Retrieval-AugmentedSurvey}. However, with the advent of Large Language Models (LLMs) and their enhanced semantic understanding capabilities, we propose broadening this definition to encompass more sophisticated transformations of documents into passages.

Different document types present distinct chunking challenges. Academic literature typically contains long sentences with a hierarchical structure, while computer log files often consist of brief, independent entries. Moreover, documents frequently contain mixed formats, including tables, lists, and figures, which traditional chunking methods handle poorly. These variations motivate the need for a more flexible framework.

We propose a formal definition of chunking to accommodate this broader perspective:


\begin{definition}[Chunking]
\label{eq:chunking_def}
A transformation $T: D \rightarrow \mathcal{P}$ that maps a document $D$ to a set of passages $\mathcal{P} = \{p_1, p_2, \dots\}$ such that:
1) The semantic information of $D$ is preserved in $\mathcal{P}$.
2) The transformation may be reversible but is not required to be.
\end{definition}


Although recent advances in RAG, such as GraphRAG \cite{Edge2024FromSummarization} and KAG \cite{Liang2024KAG:Generation}, demonstrate the potential to integrate knowledge graph construction with chunking, we focus our evaluation methodology specifically on the passage creation process to maintain clear scope.

\subsection{Principals of Chunking}
\label{subsec:principals_of_chunking}




We intend to formalise principles for chunking based on the workings of embedding models and RAG systems. These principles then serve as a starting point for the evaluation of passages. As pointed out by Wu~et.~al.~\cite{Wu2024Retrieval-AugmentedSurvey}, each passage should convey a single core concept. When multiple concepts are present in a passage, the resulting vector representation can be noisy. 
Thus, the underlying concepts are not well represented in the embedding space.
The first principle of chunking is therefore:


\newtheorem{principle}{Principle}
\begin{principle}
    \label{principle:chunk_1}
    Passages should convey one core concept.
\end{principle}

Current RAG architectures evaluate semantic similarity primarily through query-passage relationships without comprehensively accounting for interpassage semantic dependencies. While an initial retrieved passage may contain query-relevant information, its complete interpretation often depends on contextual information stored in non-retrieved passages, leading to responses generated on incomplete information. Recent work by Zhong~et~al.~\cite{Zhong2024Mix-of-Granularity:Generation}  addresses this limitation through graph-based architectures that capture passage-passage semantic relations. However, this approach introduces additional computational complexity. A more fundamental solution lies in ensuring the semantic independence of the passage during the chunking process, making the semantic information an intrinsic property of the passages.

\begin{principle}
    \label{principle:chunk_2}
    Passages should be semantically independent.
\end{principle}

Building upon the definitions of chunking (Definition~\ref{eq:chunking_def}), the third principle addresses the preservation of semantic information during the chunking process. While semantic preservation may seem straightforward, its implications for system performance are profound. Information loss at the chunking stage affects all later stages of RAG pipelines, including indexing, retrieval, and generation. Potentially compromising the effectiveness of downstream tasks and the reliability of generated responses. Therefore, we propose our third principle:

\begin{principle}
\label{principle:chunk_3}
The passage set should convey all the semantic information found in the source document.
\end{principle}

The provided principles are not absolute, and on several occasions, it will not be possible to fulfill all of them due to the structure of the underlying data. Occasionally, the principles will be conflicting such as having content with significant contradictions and exceptions, or with complex information relations, will struggle to fulfill the provided principles. However, any universal chunking method should strive to meet the principles, and the principles should form the basis for a universal chunking metric.




\section{Related Work}
\label{sec:related_work}

This section provides an overview of previous work to support the need for a chunking evaluation methodology and background knowledge on chunking methods and NLP metrics.

\subsection{LLM Sensitivity to Context}
The effectiveness of LLM-based systems depends significantly on how retrieved information is structured and presented in the prompt. Cuconasu \& Trappolini~\cite{Cuconasu2024TheSystems} demonstrated that distracting passages, while topically related, can reduce answer accuracy from $56.42\%$ to $17.95\%$ for a Llama2 model~\cite{Touvron2023LlamaModels}. Their work established that passage ordering within prompts is crucial, with information proximity to the question strongly correlating with its influence on the response. Wu \& Xie~\cite{Wu2024HowModels} further quantified this sensitivity, showing that exposure to irrelevant data can flip correct responses to incorrect ones in $15\%$ of cases event for strong models like GPT-4~\cite{OpenAI2023GPT-4Report}. These findings underscore the importance of sophisticated chunking strategies that minimize noise while preserving semantic coherence in the retrieved passages.

\subsection{Related Fields}
\label{sub:related_metrics_and_fields}
Chunking, as defined in this article (Definition~\ref{eq:chunking_def}), shares characteristics with other Natural Language Generation (NLG) tasks, particularly summarization and Machine Translation (MT). Like summarization, chunking must preserve semantic independence between segments (principle \ref{principle:chunk_2}), while like MT, it must maintain contextual meaning (principle \ref{principle:chunk_3}).
Traditional evaluation metrics for these NLG tasks, such as BLEU \cite{Papineni2002BLEU:Translation} and ROUGE \cite{Chin-YewLin2004ROUGE:Chin-Yew}, rely on lexical overlap between generated outputs and reference responses. However, these metrics inadequately capture semantic relationships crucial for chunking evaluation. While regression-based metrics like BLEURT and RUSE leverage pre-trained language models for improved semantic understanding, they require additional annotated datasets and face challenges with annotator bias \cite{Lee2023ATranslation}. 

\begin{figure*}[h!] 
\centering 
\includegraphics[width=1\linewidth]{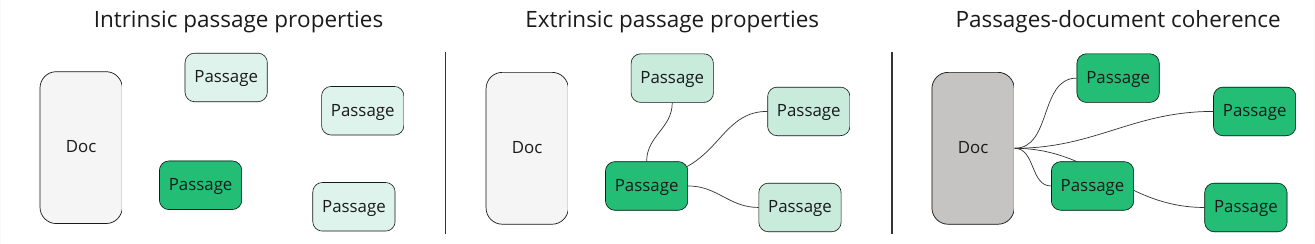} \caption{The three levels of holistic chunking evaluation. Left: The intrinsic passage properties are concerned with the information isolated within a passage. Middle: Extrinsic passage properties relate to how passages impact each other. Right: Passages-document coherence focuses on how well the passage set represents the original document.} 
\label{fig:holistic_evaluation} 
\end{figure*}

\subsection{Chunking Methods} There are several developed chunking methods of various complexity \cite{FiveLevels}, which can be roughly categorized into general-purpose methods and data-specific methods. This section focuses on general-purpose methods that maintain broad applicability across different types of documents and retrieval tasks.
\paragraph{Fixed-size Chunking} The most fundamental approach, fixed-size chunking divides text into segments of predetermined character length. This method uses two hyperparameters: \textit{passage size} 
and \textit{overlap size} 
. While effective for uniform text, this approach shows limitations with structured documents.
\paragraph{Recursive Character Chunking} This method splits text recursively on delimiter patterns, following a priority order. Commonly: double newlines, single newlines, periods, and whitespace. The process continues until all passages are below a specified maximum size, making it particularly suitable for documents with clear hierarchical organization.
\paragraph{Semantic Chunking} Semantic chunking employs embedding models to create semantically cohesive segments. The process begins with sentence-level segmentation, followed by merging segments based on their embedding similarity. Although this approach better preserves semantic relationships, its performance depends significantly on the chosen embedding model and does not take into account semantic relations that are further apart.

\section{Method} \label{sec:method} 


Building on the definition of established principles on chunking we introduced in this paper, we propose a three-level evaluation approach that examines passages at intrinsic, extrinsic, and set levels, as illustrated in Figure \ref{fig:holistic_evaluation}. This approach forms the foundation for Holistic Passage Evaluation (HOPE), a domain-agnostic metric that requires no annotated data or human intervention.
The evaluation methodology directly maps to the three fundamental principles of chunking. At the intrinsic level, we evaluate concept unity (principle~\ref{principle:chunk_1}), which measures how well individual passages maintain coherent information boundaries. The extrinsic level addresses external semantic dependence (principle~\ref{principle:chunk_2}), quantifying the independence between passages. Finally, at the set level, we assess collective information preservation (principle~\ref{principle:chunk_3}), which measures how well the complete set of passages preserves the document's original information. HOPE aggregates these three properties through arithmetic mean to produce an overall score, the HOPE metric. This design enables detailed analysis of a chunking method's strengths and weaknesses across all essential dimensions. The following subsections detail our approach to quantifying each property and their integration into the unified HOPE metric.

This part of the method utilizes LLMs as a part of several algorithms. We therefore operate with the notation $$LLM(a_1, a_2, \dots, a_n) \rightarrow b$$, where $a_1, a_2, \dots, a_n$ are text elements that are part of the input prompt, and $b$ is the LLMs text response.


\subsection{Semantic Comparison}
\label{sub:semantic_comparison}

\begin{figure*}[h]
    \centering
    \includegraphics[width=0.6\linewidth]{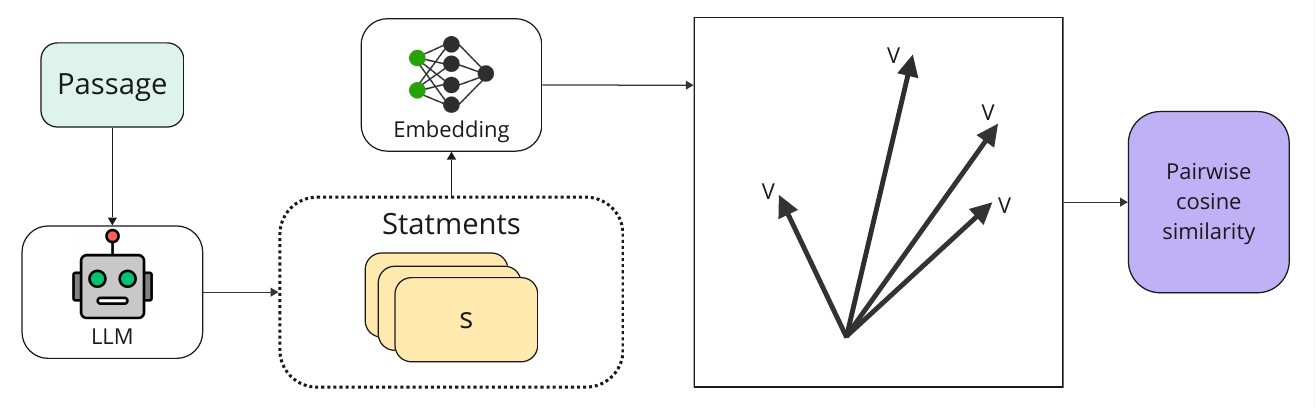}
    \caption{Calculating the concept unity $\Bar{\zeta}_{con}$: The passage in green is forwarded to an LLM to produce a set of statements $\mathcal{S}$. The statements are then transformed into vector representations using an embedding model. The pairwise cosine similarity between the vectors is calculated.}
    \label{fig:score_1_consept}
\end{figure*}

An essential component of HOPE is to compare the semantics of two or more texts. Most approaches in the literature on semantic comparison typically fall into two categories: dedicated pairwise text similarity models, such as BERT~\cite{Devlin2019BERT:Understanding}, and embedding-based methods that measure distances between dense vector representations~\cite{Guo2022SemanticReview}.
HOPE implements an embedding-based approach, which offers significant computational advantages. While pairwise comparison models require processing each text pair combination, embedding-based methods need to process each text segment only once, resulting in $O(n)$ time complexity versus $O(n^2)$ for pairwise comparisons. Our implementation utilizes Qwen~2.5~\cite{Li2023TowardsLearning}, a state-of-the-art embedding model.
The semantic comparison process consists of two steps, as shown in equation~\ref{eq:semantic_comparison}. First, an embedding model $\sigma(\cdot)$ encodes input text into dense $N$-dimensional vectors that capture semantic properties~\cite{Patil2023ANLP}. Then, the semantic similarity between two text snippets $x_1$ and $x_2$ is computed using cosine similarity $\theta(\cdot)$, which measures the angle between their vector representations $\vec{x}_1$ and $\vec{x}_2$. 
\begin{equation}
    \label{eq:semantic_comparison}
    \begin{split}
        \sigma : x &\rightarrow \vec{x} \in \mathbb{R}^N \\
        \theta(x_1, x_2) &= \frac{\sigma(x_1)\cdot\sigma(x_2)}{||\sigma(x_1)|| \cdot ||\sigma(x_2)||} \in [-1, 1]
    \end{split}
\end{equation}

While this approach efficiently captures semantic relationships, it is important to note that embedding-based methods may occasionally miss nuanced semantic differences that dedicated similarity models can detect.





\subsection{Concept Unity}
\label{sub:conceptualism}

Derived from principle~\ref{principle:chunk_1} of chunking, ``concept unity" states that a passage should have one distinct semantic meaning. While evaluating semantics presents challenges due to the complexity of natural language, LLMs offer a robust solution for assessing concept unity in text segments, demonstrating strong semantic understanding while providing consistency and scalability \cite{Zhao2023AModels}.

Inspired by claim decomposition \cite{Wanner2024ADecomposition, Gunjal2024MolecularVerification}, we propose using an LLM with a non-zero temperature parameter to generate a set of statements $\mathcal{S} = \{s_1, s_2, \dots \}$ related to a selected passage $p^* \in \mathcal{P}$. The temperature parameter introduces controlled variability in statement generation, enhancing concept coverage. If the passage contains a single concept, all generated statements should exhibit high semantic similarity, which we measure using the cosine similarity of vector embeddings as shown in equation~\ref{eq:conceptualism_score} and illustrated in Figure~\ref{fig:score_1_consept}.

\begin{equation}
    \label{eq:conceptualism_score}
    \begin{split}
        LLM(p^*) \rightarrow& \mathcal{S} = \{s_1, s_2, \dots\} \\
        \Bar{\zeta}_{con} =& \frac{1}{| \mathcal{S} |^2} \sum_{s_j \in \mathcal{S}} \sum_{s_i \in \mathcal{S}} \theta(s_i, s_j) 
    \end{split}
\end{equation}

The Concept Unity $\zeta_{con}$ is bounded by the cosine similarity interval $[-1, 1]$. A value close to $1$ indicates a homogeneous set of concepts, optimal for principle~\ref{principle:chunk_1} compliance. Negative values indicate opposing concepts, while values near $0$ indicate distantly related concepts, both undesirable. We set negative values to $0$, bounding $\Bar{\zeta}_{con}$ to $[0, 1]$. An aggregated value $\zeta_{con}$ for the entire collection of passages $\mathcal{P}$ is computed as shown in equation~\ref{eq:agregated_sem_score}.

\begin{equation}
    \label{eq:agregated_con_score}
    \begin{split}
        \zeta_{con} &= \frac{1}{| \mathcal{P} |} \sum_{p \in \mathcal{P}} \Bar{\zeta}_{con}(p, \mathcal{P})
    \end{split}
\end{equation}


To illustrate the practical application of concept unity measurement, consider the following examples demonstrating low and high concept unity:

\begin{tcolorbox}[colback=blue!5!white, colframe=black!80!red, title=Low Concept Unity]
\begin{verbatim}
Joe walked to the store to buy some veggies for 
dinner, but the fact that the boss had purchased 
a pink Mercedes was something he did not understand.
\end{verbatim}
\end{tcolorbox}

\begin{tcolorbox}[colback=blue!5!white, colframe=black!80!red, title=High Concept Unity]
\begin{verbatim}
Joe walked to the store to buy some veggies for 
dinner, but he could not decide between green, 
red or yellow bell pepper.
\end{verbatim}
\end{tcolorbox}

The first example contains two concepts: Joe at the grocery store, and Joe wondering why his boss had bought a pink Mercedes. These two concepts are not related, and by \hyperref[principle:chunk_1]{the first principle of chunking} should not be in the same passage, thus a low Concept Unity. The second example only contains information about Joe at the store, hence only a single concept, and high Concept Unity.

\subsection{Semantic Independence}
\label{sub:semantic_dependence}

According to our definition of principle~\ref{principle:chunk_2} of chunking, passages should be semantically independent. This independence implies that a passage's interpretation should remain consistent regardless of other passages present in the context. When an LLM performs open-book Q\&A, the interpretation of any given passage should remain stable, unaffected by the presence of other passages.

We propose a method for evaluating the semantic independence of a passage 
$p^*$ relative to the remaining passages $\mathcal{P}$. First, an LLM with a non-zero temperature parameter generates a set of questions $\mathcal{Q} = {q_1, q_2, \dots}$ based on the information from $p^*$. By design, the information in $p^*$ should be sufficient to answer these questions. When passages are semantically independent, the LLM's response $a^*$ should remain consistent even when a subset $\mathcal{P}_q \subset \mathcal{P}$ is present, as shown in equation~\ref{eq:semantic_dependency}.

\begin{equation}
    \label{eq:semantic_dependency}
    \begin{split}
        LLM(q, p^*) &\rightarrow a^*  \\
        LLM(q, p^*, \mathcal{P}_q) &\rightarrow a    \\
        a^* &\simeq a
    \end{split}
\end{equation}

The subset $\mathcal{P}_q$ comprises the top-k passages from $\mathcal{P}$ that maximize semantic similarity to the question $q$, where k is set to 3 in our implementation. The two LLM configurations from equation~\ref{eq:semantic_dependency} each generate a set of answers $\mathcal{A} = \{a_1, a_2, \dots\}$ and $\mathcal{A}^* = \{a^*_1, a^*_2, \dots\}$ to the questions in $\mathcal{Q}$. These answers are compared pairwise using cosine similarity $\theta(a^*, a)$, as shown in equation~\ref{eq:semantic_dependency_score}.

\begin{equation}
    \label{eq:semantic_dependency_score}
    \begin{split}
        LLM(q, p^*), \forall q \in \mathcal{Q} &\rightarrow \mathcal{A}^* = \{a^*_1, a^*_2, \dots\}  \\
        LLM(q, p^*, \mathcal{C}_q), \forall q \in \mathcal{Q} &\rightarrow \mathcal{A} = \{a_1, a_2, \dots\} \\
        \Bar{\zeta}_{sem} &= \frac{1}{|\mathcal{A}|} \sum_{a \in \mathcal{A}, a^* \in \mathcal{A}^*} \theta(a^*, a) 
    \end{split}
\end{equation}

\begin{figure*}[h]
    \centering
    \includegraphics[width=0.6\linewidth]{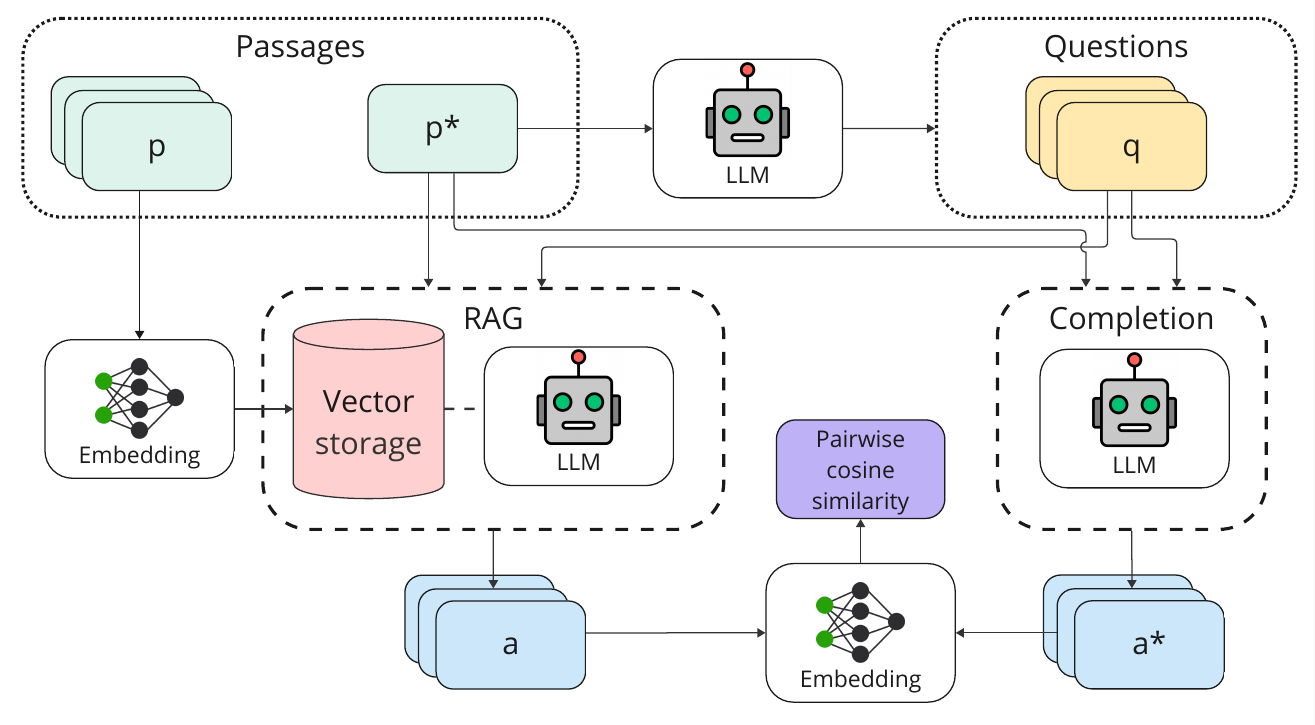}
    \caption{Calculation of the semantic independence $\Bar{\zeta}_{sem}$: A selected passage $p^*$ is used to construct a set of questions $Q$. The questions are then answered by two LLMs: one that can access all passages $\mathcal{P}$ (RAG) and one that can only access the focus passage $p^*$ (Completion). The two LLMs produce the answers $\mathcal{A}$ and $\mathcal{A}^*$, which are then compared using an embedding model and pairwise cosine similarity.}
    \label{fig:score_2_sematics}
\end{figure*}

$\Bar{\zeta}_{sem}$ is bounded by the cosine similarity interval $[-1, 1]$. A value of $1$ indicates complete semantic independence, while values near $0$ suggest strong semantic dependence. Values approaching $-1$ indicate inverted semantics, which can potentially mislead LLMs~\cite{Wu2024HowModels}. Since both semantic alteration and inversion are undesirable, we cap $\Bar{\zeta}_{sem}$ to the interval $[0, 1]$ by setting negative values to zero. Figure~\ref{fig:score_2_sematics} illustrates the calculation process of $\Bar{\zeta}_{sem}$. An aggregated value $\zeta_{sem}$ for the entire collection $\mathcal{P}$ is computed as the average of individual passage scores, as shown in equation~\ref{eq:agregated_sem_score}.

\begin{equation}
    \label{eq:agregated_sem_score}
    \begin{split}
         \zeta_{sem} &= \frac{1}{| \mathcal{P} |} \sum_{p \in \mathcal{P}} \Bar{\zeta}_{sem}(p, \mathcal{P})
    \end{split}
\end{equation}

To illustrate semantic independence, consider the following examples demonstrating low and high semantic independence:

\begin{tcolorbox}[colback=blue!5!white, colframe=black!80!red, title=Low Semantic Independence]
\begin{verbatim}
#Passage 1
Joe is a professional driver. He is therefore 
allowed to drive over the speed limit. 

#Passage 2
Exceptions from the speed limits 
only apply during the daytime.
\end{verbatim}
\end{tcolorbox}

\begin{tcolorbox}[colback=blue!5!white, colframe=black!80!red, title=High Semantic Independence]
\begin{verbatim}
#Passage 1
Joe is a professional driver. He is therefore 
allowed to drive over the speed 
limit during the daytime. 

#Passage 2
Exceptions from the speed limits 
only apply during the daytime. 
\end{verbatim}
\end{tcolorbox}

In the first example, semantic independence is low because passage 2 contains critical information that modifies the interpretation of passage 1, namely the exception related to daytime. The second example achieves high semantic independence by incorporating all related contexts within passage 1.


\subsection{Collective Information Preservation}
\label{sub:information_preservation}

\begin{figure*}[h]
    \centering
    \includegraphics[width=0.6\linewidth]{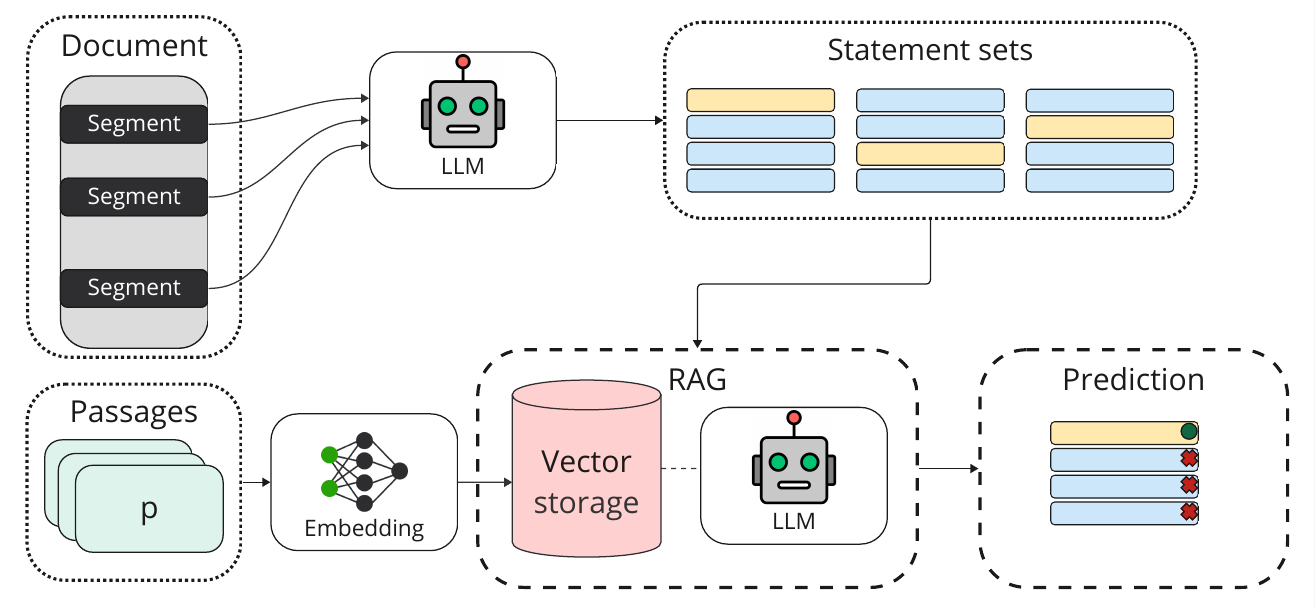}
    \caption{Calculation of the information preservation $\zeta_{inf}$: Three-sentence segments are randomly sampled from the original document. An LLM uses the segments to construct statement quadruplets, where one is verifiable true and three are plausible but false. A secondary LLM is then tasked with truth discrimination by analyzing the statements against contextually relevant passages.}
    \label{fig:score_3_information}
\end{figure*}

The preservation of information during document chunking represents a critical requirement as established by principle~\ref{principle:chunk_3}. While previous characteristics (subsection~\ref{sub:conceptualism} and \ref{sub:semantic_dependence}) evaluate passages from the view of single passages, quantifying information loss demands simultaneous analysis across all passages and the source document. In an ideal scenario, there would exist a comprehensive method $I(\cdot)$ capable of extracting all atomic facts $f \in \mathcal{F}$ from text, where $\mathcal{F}$ encompasses the complete space of factual statements expressible in natural language. This method would extract every atomic fact from the original document $D$ and verify their preservation within the collection of passages $\mathcal{P}$, as formalized in equation~\ref{eq:infomration_subset}.

\begin{equation}
    \label{eq:infomration_subset}
    \begin{split}
        I : x &\rightarrow \mathcal{F}_x = \{ f_1, f_2, \dots \} \subseteq \mathcal{F} \\
        \mathcal{F}_D &\subseteq \mathcal{F}_\mathcal{P}
    \end{split}
\end{equation}



However, the implementation of an ideal method $I(\cdot)$ proves impractical for extensive texts due to the inherent complexity of natural language. This complexity makes the complete enumeration of atomic facts virtually impossible. To address this challenge, we propose leveraging a subset of atomic facts $\hat{\mathcal{F}}_D \subseteq \mathcal{F}_D$ as a robust approximation. This subset can be derived through a non-ideal yet practical function $\hat{I}(\cdot)$ that, while not exhaustive, effectively captures key factual information.

\begin{equation}
    \label{eq:information_estimation_subset}
    \begin{split}
        \hat{I} : x &\rightarrow \hat{\mathcal{F}_x} = \{ f_1, f_2, \dots \} \subset \mathcal{F}_x \\ 
        \hat{\mathcal{F}}_D \subseteq \mathcal{F}_D \subseteq \mathcal{F}_\mathcal{P} &\Rightarrow \hat{\mathcal{F}}_D \subseteq \mathcal{F}_\mathcal{P}
    \end{split}
\end{equation}


As equation~\ref{eq:information_estimation_subset} delineates, the subset of atomic facts $\hat{\mathcal{F}}_D$ should be fully contained within the space of the chunked sections $\mathcal{F}_C$. The degree of overlap between these sets serves as a decisive indicator of information preservation across the chunking process.


LLMs emerge as an optimal implementation choice for $\hat{I}(\cdot)$, given their exceptional capabilities in semantic comprehension and nuanced fact extraction from natural language \cite{Zhao2023AModels}. Their demonstrated proficiency in identifying and validating factual relationships within text positions them as ideal candidates for approximating the theoretical information extraction function $I(\cdot)$.


Our method involves sampling document segments $d$ comprising three consecutive sentences from the original document $D$ using a uniform probability distribution $U(\cdot)$. While this localized sampling approach effectively captures information within sentence triplets, we acknowledge its current limitation in detecting complex relationships across longer distances. Each sampled segment undergoes LLM processing to generate quadruples containing one true and three false statements ${s_t, s_{f1}, s_{f2}, s_{f3}} = q$.
The system leverages cosine similarity matching to retrieve relevant passages $\mathcal{P}_{s_t} \subset \mathcal{P}$ from a vector database based on the true statement $s_t$. These retrieved passages are then analyzed by an LLM tasked with identifying the true statement among the quadruple. To ensure consistent evaluation, the LLM must provide a definitive prediction even under uncertainty. This comprehensive approach is formalized in equation~\ref{eq:collective_information_preservation} and illustrated in Figure~\ref{fig:score_3_information}.


\begin{equation}
    \label{eq:collective_information_preservation}
    \begin{split}
        d &\sim U(D) \\
        LLM(d) &\rightarrow \{s_t, s_{f1}, s_{f2}, s_{f3}\} = s \\
        LLM(s, \mathcal{P}_{s_t}) &\rightarrow a \\
        \mathcal{A} &= \{a_1, a_2, \dots\} \\
        \mathcal{S} &= \{s_1, s_2, \dots\} \\
        \zeta_{inf} &= \frac{1}{| \mathcal{S} |} \sum_{a \in \mathcal{A}, s \in \mathcal{S}} \begin{cases}
            1 & \text{if $a = s_t$} \\
            0 & \text{if $a \neq s_t$}
        \end{cases} 
    \end{split}
\end{equation}

Our method employs a binary scoring mechanism for $\zeta_{inf}$, prioritizing clarity and interpretability in measuring information preservation. While more nuanced scoring approaches might capture partial information retention, they would introduce unnecessary complexity into the evaluation methodology.


To demonstrate the practical efficacy of our information preservation metric, we present two carefully constructed examples that illustrate the nuanced ways in which chunking strategies can impact information retention. These examples illustrate how subtle variations can produce measurably different outcomes in terms of preserving critical information.


\begin{tcolorbox}[colback=blue!5!white, colframe=black!80!red, title=Low Information Preservation]
\begin{verbatim}
#Document 
Joe is a professional jogger, therefore he can 
finish a marathon in less than 3 hours.

#Passage 1
Joe is a professional jogger.

#Passage 2
Joe can finish a marathon in an impressive time.
\end{verbatim}
\end{tcolorbox}

The first example illustrates a significant degradation in information fidelity, where precision in the information is lost ("less than 3 hours" $\rightarrow$ "impressive time"). 

\begin{tcolorbox}[colback=blue!5!white, colframe=black!80!red, title=High Information Preservation]
\begin{verbatim}
#Document 
Joe is a professional jogger, therefore he can 
finish a marathon in less than 3 hours.

#Passage 1
Joe is a professional jogger.

#Passage 2
Joe can finish a marathon in less than 3 hours.
\end{verbatim}
\end{tcolorbox}

\begin{figure*}[h!]
    \centering
    \includegraphics[width=0.9\linewidth]{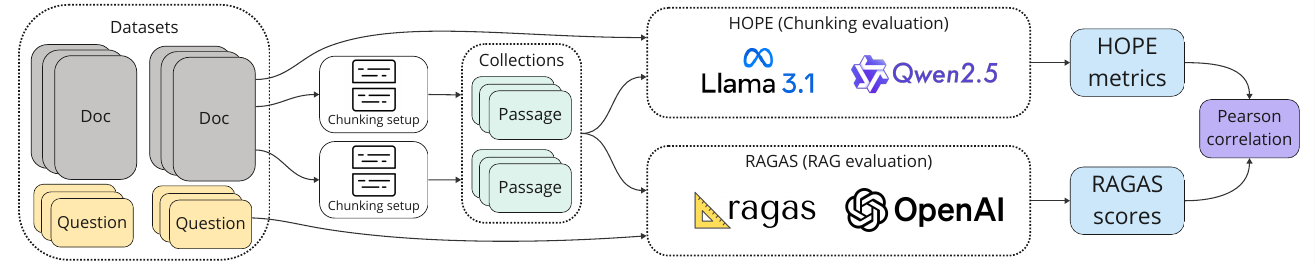}
    \caption{Documents and questions from diverse domains are processed through eight chunking configurations (four fixed-size, two recursive, and two semantic chunking variants). For each resulting passage collection, both the HOPE metric and RAGAS scores are computed. The setup then analyzes correlation between HOPE and RAGAS scores across all configurations and datasets.}
    \label{fig:chunkbench}
\end{figure*}

In contrast, the second example, the chunking strategy maintains the integrity of all atomic facts present in the original document. This faithful retention of information enables complete reconstruction of the original semantic content, resulting in a superior $\zeta_{inf}$ score.

\subsection{HOPE Metric}

A formal metric rooted in the three principles of chunking (section \ref{subsec:principals_of_chunking}), irrespective of domain, and with no need for human annotations, is needed so that more clever chunking methods can be developed and that chunking can be applied more intelligent in real-world RAG systems. 

The HOPE metric covers all three levels of passage evaluation (section \ref{sec:method}) to provide a holistic assessment of chunking quality. HOPE maps to the interval $[0, 1]$, where 1 represents ideal chunking, and 0 indicates complete chunking failure. The metric is defined as a normalized linear combination of concept unity (subsection~\ref{sub:conceptualism}), semantic independence (subsection~\ref{sub:semantic_dependence}), and information preservation (subsection~\ref{sub:information_preservation}), as shown in equation~\ref{eq:HOPE_def}.


\begin{equation}
    \label{eq:HOPE_def}
    \begin{split}
        HOPE &= \frac{1}{3} (\zeta_{inf} +  \zeta_{sem} +  \zeta_{con})
    \end{split}
\end{equation}


The implementation of these sub-metrics relies on synthetic data generation from LLMs, which presents specific technical challenges. For robust metric calculation, the generated synthetic questions and answers must satisfy two critical properties: faithfulness and diversity \cite{Long2024OnSurvey}. Insufficient faithfulness results in off-topic or inaccurate questions and statements, potentially based on hallucinated facts, violating the fundamental relationship established in equation~\ref{eq:infomration_subset}. Limited diversity in the generated content leads to inadequate representation of the underlying data, constraining the effectiveness of $\zeta_{con}$ and $\zeta_{inf}$ to a subset of the available information. The broader implications and challenges of utilizing language models for these evaluations are addressed in section \ref{sec:discussion}.



\section{Experiment Methodology and Results}
\label{sec:results}


We evaluate the effect of HOPE by first collecting a dataset consisting of a diverse set of document types and generating related questions. The documents serve as a knowledge base that we index eight times using distinct chunking setups. For each indexing, both the HOPE metric and the RAG performance indicators are calculated on a corpus level using independent queries. The variance in RAG performance indicators is caused by the effectiveness of the different chunking setups, which is what HOPE intends to model. Thus we will look at correlations between the HOPE metric and the RAG performance metrics to tell how well HOPE can model the effectiveness of chunking.



\subsection{Dataset}
Existing benchmarks like BEIR~\cite{Thakur2021BEIR:Models} and MTEB~\cite{Muennighoff2023MTEB:Benchmark} focus on retriever evaluation, and benchmarks like MMLU~\cite{Hendrycks2021MeasuringUnderstanding} target generator evaluation. Passage quality affects both retrieval and generation, thus our dataset needs to address end-to-end RAG system performance. Also, HOPE intends to be domain-agnostic, thus empirical tests must be carried out on documents from different domains and with dissimilar document structures. Following the approach of NQ-open~\cite{Lee2020LatentAnswering}, we created seven domain-specific Q\&A datasets. For each domain, we collected a set of documents and generated 100 open-book questions using the RAGAS framework. Table~\ref{tab:data_overview} presents our diverse document collection, ranging from highly structured technical manuals to more narrative-driven debate transcripts.


\raggedbottom
\begin{table}[htbp]
    \begin{tabularx}{\linewidth}{p{0.3\linewidth} p{0.2\linewidth} p{0.4\linewidth}}
        \toprule
        
         Data type          & Documents &  Source\\
        
        \midrule
         Newspapers         & 20 & AP, Aljazeera \& BBC\\
         Academic articles  & 12 & Arxiv\\
         Wiki articles      & 54 & Wikipedia\\
         Technical manuals  & 10 & VW Car Owners manual\\
         Debates            & 5  & Pile of Law \cite{hendersonkrass2022pileoflaw} \\
         Terms of service   & 10 & Pile of Law \cite{hendersonkrass2022pileoflaw}\\
         Medical            & 20 & The {COVID-19} Open Research Dataset \cite{wang-etal-2020-cord}\\    
         \bottomrule
         Total: & 131  & \\
         \bottomrule
    \end{tabularx}
    \caption{An overview of the datasets used for evaluating the HOPE metric.}
    \label{tab:data_overview}
\end{table}

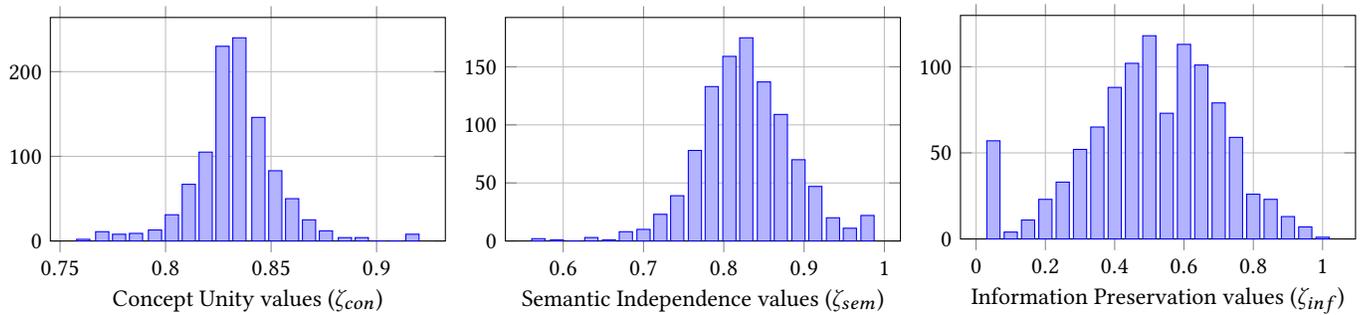
\begin{figure*}[h!]
    \centering
    \begin{minipage}{0.32\textwidth}
        \centering
        \begin{tikzpicture}[scale=1]
            \begin{axis}[
                ybar,
                bar width=5,
                xlabel={Concept Unity values ($\zeta_{con}$)},
                ymin=0,
                xtick={0.7, 0.75, 0.8, 0.85, 0.9, 0.95, 1.0},
                grid=major,
                width=\linewidth*1.2,
                height=\linewidth*0.8
            ]
                \addplot coordinates {
                    (0.761, 2)
                    (0.77, 11)
                    (0.778, 8)
                    (0.786, 9)
                    (0.795, 13)
                    (0.803, 31)
                    (0.811, 67)
                    (0.819, 105)
                    (0.827, 230)
                    (0.835, 240)
                    (0.844, 146)
                    (0.852, 83)
                    (0.86, 50)
                    (0.868, 25)
                    (0.876, 12)
                    (0.885, 4)
                    (0.893, 4)
                    (0.901, 0)
                    (0.909, 0)
                    (0.917, 8)
                };
            \end{axis}
        \end{tikzpicture}
    \end{minipage}\hfill
    \begin{minipage}{0.32\textwidth}
        \centering
        \begin{tikzpicture}[scale=1]
            \begin{axis}[
                ybar,
                bar width=5,
                xlabel={Semantic Independence values ($\zeta_{sem}$)},
                ymin=0,
                xtick={0.5, 0.6, 0.7, 0.8, 0.9, 1.0},
                grid=major,
                width=\linewidth*1.2,
                height=\linewidth*0.8
            ]
                \addplot coordinates {
                    (0.569, 2)
                    (0.592, 1)
                    (0.614, 0)
                    (0.635, 3)
                    (0.657, 1)
                    (0.678, 8)
                    (0.7, 10)
                    (0.721, 23)
                    (0.742, 39)
                    (0.764, 78)
                    (0.785, 133)
                    (0.807, 159)
                    (0.828, 175)
                    (0.85, 137)
                    (0.871, 109)
                    (0.893, 70)
                    (0.914, 47)
                    (0.936, 20)
                    (0.957, 11)
                    (0.979, 22)
                };
            \end{axis}
        \end{tikzpicture}
    \end{minipage}\hfill
    \begin{minipage}{0.32\textwidth}
        \centering
        \begin{tikzpicture}[scale=1]
            \begin{axis}[
                ybar,
                bar width=5,
                xlabel={Information Preservation values ($\zeta_{inf}$)},
                ymin=0,
                xtick={0.0, 0.2, 0.4, 0.6, 0.8, 1.0},
                grid=major,
                width=\linewidth*1.2,
                height=\linewidth*0.8
            ]
                \addplot coordinates {
                    (0.05, 57)
                    (0.1, 4)
                    (0.15, 11)
                    (0.2, 23)
                    (0.25, 33)
                    (0.3, 52)
                    (0.35, 65)
                    (0.4, 88)
                    (0.45, 102)
                    (0.5, 118)
                    (0.55, 73)
                    (0.6, 113)
                    (0.65, 101)
                    (0.7, 79)
                    (0.75, 59)
                    (0.8, 26)
                    (0.85, 23)
                    (0.9, 13)
                    (0.95, 7)
                    (1.0, 1)
                };
            \end{axis}
        \end{tikzpicture}
    \end{minipage}
    \caption{Distributions of the HOPE values for all 1048 combinations of documents and chunking methods.}
    \label{fig:dist_hope_scores}
\end{figure*}

\subsection{RAG Evaluation}
RAG performance evaluation is composed of a diverse set of metrics to isolate and test specific aspects of the RAG system. We utilize RAGAS~\cite{Es2024RAGAS:Generation}, which has developed several LLM-based metrics for evaluating both the final response and the information retrieval components.

To explore how the HOPE metrics relate to RAG performance, we selected four metrics that assess both generation quality and retrieval effectiveness. Three metrics analyze information-related characteristics of the response, including semantics, factuality, and structural aspects, while the fourth metric evaluates retrieval quality. The metrics are implemented using the RAGAS python library\footnote{\url{https://docs.ragas.io/en/stable/}}:



\begin{itemize}
    \item \textbf{Answer Correctness (AC)} Measures answer correctness compared to ground truth as a combination of factuality and semantic similarity. 
    \item \textbf{Response Relevancy (RR)} Scores the relevancy of the answer according to the given question. Answers with incomplete, redundant, or unnecessary information are penalized. 
    \item \textbf{Factual Correctness (FC)} Uses claim decomposition and natural language inference (NLI) to verify the claims made in the responses against reference texts.
    \item \textbf{Context Recall (CR)} Estimates context recall by estimating TP and FN using annotated answers and retrieved context. 
\end{itemize}

These metrics allow us to evaluate both the effectiveness of information retrieval and the quality of fact preservation across different chunking strategies, directly addressing principles \ref{principle:chunk_2} and \ref{principle:chunk_3} of chunking.

\subsection{Results}

Eight different constellations of chunking methods were evaluated to test the versatility of HOPE. These include four fixed-size, two recursive, and two semantic chunking methods. The fixed-size and recursive approaches operate with either large passages (2000 characters) or small passages (500 characters), while the fixed-size method additionally varies the overlap size (500 characters and 125 characters). For semantic chunking, we employed two embedding models: OpenAI's text-embedding-ada-002~\cite{Neelakantan2022TextPre-Training} and Alibaba's Qwen2 model~\cite{Li2023TowardsLearning}.

We calculated the RAG performance indicators using OpenAI's GPT-4o-mini~\cite{OpenAI2023GPT-4Report} and text-embedding-ada-002~\cite{Neelakantan2022TextPre-Training}, while the HOPE metrics\footnote{The link to the python implementation of HOPE is removed during the review phase to preserve anonymity, but will be available in the published versions.} were computed using NVIDIA's Nemotron-70B~\cite{Wang2024HelpSteer2-Preference:Preferences} (based on Llama 3.1 70B~\cite{Dubey2024TheModels}) and Alibaba's Qwen2.5~\cite{Li2023TowardsLearning}, as illustrated in Figure~\ref{fig:chunkbench}. The distributions of the score for each of the three characteristics across all samples are displayed in Figure~\ref{fig:dist_hope_scores}, with Pearson correlations between RAG and HOPE metrics presented in Table~\ref{tab:correlations}. As a baseline comparison, we calculated BLEU scores by comparing the original document with a sequential concatenation of the passages. While BLEU serves as an established metric in NLP evaluations, it provides a conservative baseline for assessing chunking quality.

\begin{table}
    \centering
    \begin{tabularx}{0.85\linewidth}{c c c c c c}
         \toprule 
                &  BLEU     & HOPE              & $\zeta_{con}$     & $\zeta_{sem}$     & $\zeta_{inf}$ \\
         \midrule
         AC     & -0.010   & 0.024              & -0.024            & \textbf{0.105}*   & 0.002 \\
         FC     & -0.015   & 0.080*             & -0.036            & \textbf{0.136}*   & 0.053 \\
         CR     & -0.011   & 0.011              & -0.103*           & \textbf{0.117}*   & -0.010 \\
         RR     & 0.027    & \textbf{0.099}*    & -0.068*           & 0.054             & 0.091* \\
         \bottomrule
    \end{tabularx}
    \caption{Pearson correlations ($\rho$) between RAG performance indicators, and HOPE-metrics and BLEU score. All values marked with an asterisk * are statistically significant with $\text{p-value} < 0.05$.}
    \label{tab:correlations}
\end{table}

\begin{figure}
    \begin{tikzpicture}
        \begin{axis}[
            width=9.5cm, height=6cm,
            xlabel={Semantic Independence values},
            xtick=data,
            xticklabel style={rotate=30, anchor=east},
            legend pos=north west,
            grid=both,
            ymin=0.1, ymax=1, 
            legend style={
                font=\small, 
                at={(0.82, 0.23)},
                anchor=west, 
                align=left
            } 
        ]
            \addplot+[
                mark=o, 
                nodes near coords, 
                point meta=explicit symbolic
            ] coordinates {
                (1, 0.531) [0.531] (2, 0.526) [0.526] (3, 0.550) [0.550] 
                (4, 0.557) [0.557] (5, 0.562) [0.562] (6, 0.546) [0.546] 
                (7, 0.582) [0.582] (8, 0.572) [0.572] (9, 0.642) [0.642] 
                (10, 0.643) [0.643]
            };
            \addlegendentry{AC}
    
            \addplot+[
                mark=o, 
                nodes near coords, 
                point meta=explicit symbolic
            ] coordinates {
                (1, 0.345) [0.345] (2, 0.343) [0.343] (3, 0.347) [0.347] 
                (4, 0.410) [0.410] (5, 0.355) [0.355] (6, 0.329) [0.329] 
                (7, 0.393) [0.393] (8, 0.380) [0.380] (9, 0.556) [0.556] 
                (10, 0.539) [0.539]
            };
            \addlegendentry{FC}
    
            \addplot+[
                mark=o, 
                nodes near coords, 
                point meta=explicit symbolic
            ] coordinates {
                (1, 0.761) [0.761] (2, 0.750) [0.750] (3, 0.783) [0.783] 
                (4, 0.809) [0.809] (5, 0.839) [0.839] (6, 0.838) [0.838] 
                (7, 0.868) [0.868] (8, 0.882) [0.882] (9, 0.873) [0.873] 
                (10, 0.834) [0.834]
            };
            \addlegendentry{CR}
    
            \addplot+[
                mark=o, 
                nodes near coords, 
                point meta=explicit symbolic
            ] coordinates {
                (1, 0.742) [0.742] (2, 0.780) [0.780] (3, 0.797) [0.797] 
                (4, 0.823) [0.823] (5, 0.773) [0.773] (6, 0.836) [0.836] 
                (7, 0.834) [0.834] (8, 0.800) [0.800] (9, 0.820) [0.820] 
                (10, 0.805) [0.805]
            };
            \addlegendentry{RR}

            \pgfplotsset{
                xtick={1, 2, 3, 4, 5, 6, 7, 8, 9, 10},
                xticklabels={
                    {0.786},
                    {0.796},
                    {0.806},
                    {0.817},
                    {0.827},
                    {0.837},
                    {0.848},
                    {0.858},
                    {0.868},
                    {0.879}
                }
            }
        \end{axis}
    \end{tikzpicture}
    \caption{RAG performance indicators plotted against the Semantic Independence values found in Figure~\ref{fig:dist_hope_scores}.}
    \label{fig:sem_ind_rag_metrics}
\end{figure}
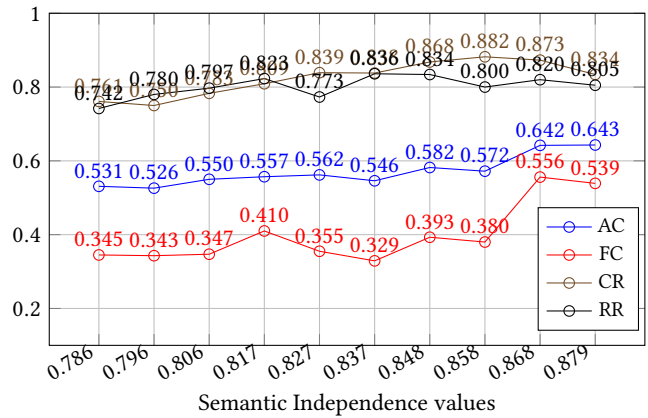

\section{Discussion}
\label{sec:discussion}

Our analysis demonstrates that lexical approaches such as BLEU fail to adequately capture chunking quality, as evidenced by their low correlations with RAG performance metrics (Table~\ref{tab:correlations}). This limitation is unsurprising given BLEU's fundamental design, which lacks semantic understanding capabilities~\cite{Lee2023ATranslation}. In contrast, HOPE exhibits significant correlations with various RAG performance metrics, particularly through its Semantic Independence component, which shows strong associations with information-based metrics including Answer Correctness and Factual Correctness. The relationship becomes particularly evident in Figure~\ref{fig:sem_ind_rag_metrics}, which illustrates consistent improvements across all RAG performance indicators as Semantic Independence values ($\zeta_{sem}$) increase. The magnitude of this effect is substantial, with performance gains of $\textbf{21.1\%}$ and $\textbf{56.2\%}$ for Answer Correctness and Factual Correctness, respectively, when comparing readings at minimum and maximum Semantic Independence values. These findings substantiate Semantic Independence as a crucial characteristic in chunking method design, lending empirical support to the second principle of chunking (Principle~\ref{principle:chunk_2}). Moreover, techniques such as decontextualization~\cite{Gunjal2024MolecularVerification, Newman2023ADocuments}, previously employed in claim decomposition, emerge as promising approaches for enhancing Semantic Independence during chunking.

\subsection{Challenging Traditional Assumptions}
A particularly intriguing finding emerges from our analysis of Concept Unity values ($\zeta_{con}$), which exhibit negative correlations with all RAG performance indicators (Table~\ref{tab:correlations}). This unexpected result challenges the validity or completeness of the first principle of chunking (Principle~\ref{principle:chunk_1}), suggesting that the conventional idea of isolating concepts within passages may be suboptimal. We propose three plausible explanations for this counterintuitive phenomenon: First, the semantic proximity of concepts may prevent the generation of noise in embedding vectors; second, embedding models may be intrinsically optimized for multi-concept passages due to their training data composition; and third, single-concept passages may suffer from reduced information density, potentially diminishing their utility during response generation. The distribution analysis of HOPE metrics (Figure~\ref{fig:dist_hope_scores}) reveals that current Concept Unity implementations generate values within a constrained range ($\approx10\%$ of total range), indicating limited diversity in LLM-generated statements—a recognized limitation in synthetic data generation~\cite{Long2024OnSurvey}. While a specialized embedding model for Concept Unity calculation could potentially expand this range, such an approach would contradict HOPE's objective of serving as an automatic metric.

\subsection{Limitations and Technical Considerations}

The implementation of HOPE as an automatic metric necessitates the integration of LLMs and embedding models, introducing variability and uncertainty, which any language model potentially inherits from biases found in its training corpus \cite{Zhao2023AModels}. An LLM's bias can lead to less diverse statements in certain areas or topics, thus directly impacting the Concept Unity and Information Preservation components. While LLM performance can be significantly enhanced through sophisticated prompt engineering~\cite{White2023AChatGPT}, our current implementation employs relatively straightforward prompts with modest in-context learning examples. More advanced prompt engineering techniques like chain-of-thought reasoning~\cite{Wei2022Chain-of-ThoughtModels} and LLM-based agent networks~\cite{Xi2023TheSurvey} could potentially improve the statement generation diversity for both Concept Unity and Information Preservation components, but this is left as future work.

\subsection{Information Preservation and Response Quality}

Our examination reveals that Information Preservation demonstrates a positive correlation with RAG performance (Table~\ref{tab:correlations}). The metric exhibits a comprehensive range of values, providing additional validation for the third principle of chunking (Principle~\ref{principle:chunk_3}). Furthermore, the strong correlation between Information Preservation and Response Relevancy suggests a fundamental relationship between information completeness and response quality. This association is theoretically consistent, as Response Relevancy inherently penalizes incomplete information — precisely the phenomenon that Information Preservation aims to quantify.

\subsection{Implications and Future Directions}

Our findings have substantial implications for the evolution of RAG system architecture. The correlation between Semantic Independence and RAG performance suggests that future system designs should prioritize this characteristic during passage optimization, but more research is needed to verify this empirically. One promising avenue involves leveraging either HOPE or its Semantic Independence component as reward functions within reinforcement learning frameworks for chunk boundary optimization. However, the computational intensity of HOPE, primarily due to multiple LLM invocations, presents significant scalability challenges that warrant further investigation.

Several critical research directions emerge from our analysis. First, the superior performance of multi-concept passages merits deeper investigation, potentially through detailed analysis. Second, the development of computationally efficient chunking strategies that optimize for Semantic Independence while maintaining Information Preservation could yield substantial improvements in RAG system performance. This could deem challenging as Semantic Independence would likely benefit from rewriting the original content, while Information Preservation would benefit from preserving the original structure to avoid "loss in translation".

\section{Conclusion}
\label{sec:conclusion}
This paper advances the understanding of document chunking in RAG systems through three main contributions. First, we introduce a methodology for characterizing chunking at three separate levels: intrinsic passage properties, extrinsic passage properties, and passages-document coherence. Second, we propose HOPE, a domain-agnostic metric that quantifies these characteristics and provides a systematic approach to evaluating chunking strategies. Third, through empirical evaluation across seven domains, we demonstrate that semantic independence between passages significantly impacts RAG performance, yielding improvements of up to $56.2\%$ in Factual Correctness and $21.1\%$ in Answer Correctness.

Our findings challenge traditional assumptions about Concept Unity within passages, revealing its minimal impact on system performance. These insights provide guidance for optimizing chunking strategies in RAG systems. Future work could explore the adaptation of the HOPE metric to specific domains and investigate its applicability in dynamic chunking scenarios. We believe our methodology lays a solid foundation for systematic improvements of chunking strategies in RAG applications.

\balance
\printbibliography




\end{document}